\documentclass[conference]{IEEEtran}
\IEEEoverridecommandlockouts
\usepackage{cite}
\usepackage{amsmath,amssymb,amsfonts}
\usepackage{algorithmic}
\usepackage{graphicx}
\usepackage{textcomp}
\usepackage{xcolor}
\def\BibTeX{{\rm B\kern-.05em{\sc i\kern-.025em b}\kern-.08em
    T\kern-.1667em\lower.7ex\hbox{E}\kern-.125emX}}

\usepackage{booktabs}
\usepackage{multirow}
\usepackage{bm}
\usepackage{mathrsfs}
\usepackage{bbding}
\usepackage[misc]{ifsym}

\begin{document}

\title{Plug-and-Play Feature Generation for Few-Shot Medical Image Classification
\thanks{*: Corresponding Authors.}
}

\author{\IEEEauthorblockN{Qianyu Guo}
\IEEEauthorblockA{\textit{School of Computer Science} \\
\textit{Fudan University}\\
Shanghai, China \\
qyguo20@fudan.edu.cn}
\and
\IEEEauthorblockN{Huifang Du}
\IEEEauthorblockA{\textit{College of Design and Innovation} \\
\textit{Tongji University}\\
Shanghai, China\\
duhuifang@tongji.edu.cn}
\and
\IEEEauthorblockN{Xing Jia}
\IEEEauthorblockA{\textit{School of Computer Science} \\
\textit{Fudan University}\\
Shanghai, China \\
xjia18@fudan.edu.cn}
\and
\IEEEauthorblockN{Shuyong Gao}
\IEEEauthorblockA{\textit{School of Computer Science} \\
\textit{Fudan University}\\
Shanghai, China \\
sygao18@fudan.edu.cn}
\and
\IEEEauthorblockN{Yan Teng}
\IEEEauthorblockA{\textit{Governance Research Center} \\
\textit{Shanghai Artificial Intelligence Laboratory}\\
Shanghai, China \\
tengyan@pjlab.org.cn}
\and
\IEEEauthorblockN{Haofen Wang\footnotesize \textsuperscript{*}}
\IEEEauthorblockA{\textit{College of Design and Innovation} \\
\textit{Tongji University}\\
Shanghai, China \\
carter.whfcarter@gmail.com}
\and
\IEEEauthorblockN{Wenqiang Zhang\footnotesize \textsuperscript{*}}
\IEEEauthorblockA{\textit{School of Computer Science} \\
\textit{Fudan University}\\
Shanghai, China \\
wqzhang@fudan.edu.cn}
}

\maketitle

\begin{abstract}
Few-shot learning (FSL) presents immense potential in enhancing model generalization and practicality for medical image classification with limited training data; however, it still faces the challenge of severe overfitting in classifier training due to distribution bias caused by the scarce training samples.
To address the issue, we propose MedMFG, a flexible and lightweight plug-and-play method designed to generate sufficient class-distinctive features from limited samples.
Specifically, MedMFG first re-represents the limited prototypes to assign higher weights for more important information features. Then, the prototypes are variationally generated into abundant effective features. Finally, the generated features and prototypes are together to train a more generalized classifier.
Experiments demonstrate that MedMFG outperforms the previous state-of-the-art methods on cross-domain benchmarks involving the transition from natural images to medical images, as well as medical images with different lesions.
Notably, our method achieves over 10\% performance improvement compared to several baselines. Fusion experiments further validate the adaptability of MedMFG, as it seamlessly integrates into various backbones and baselines, consistently yielding improvements of over 2.9\% across all results.

\end{abstract}


\section{Introduction}

 Few-shot learning (FSL)~\cite{wang2020generalizing,guo2020broader,guo2023rankdnn} has been widely adopted to address the challenges of limited annotated data and the long-tail distribution in medical image classification, as it enables rapid acquisition of new category knowledge from limited samples. In few-shot medical image classification, the feature extraction model is trained on a large number of natural or common medical images in the source domain. Subsequently, the target domain images extract prototype features and train the classifier. Nevertheless, the limited data in the target domain fails to represent the class distribution accurately. Additionally, due to the semantic gap between the source and target domains, the pre-trained model struggles to accurately capture the prototype features of the target domain data. Noisy prototypes further contribute to the errors in the category distribution. Consequently, the biased representation and scarcity of training samples result in a skewed category distribution, leading to severe overfitting during classifier training and impacting the performance of few-shot medical image classification.

Existing methods in medical few-shot learning can be broadly categorized into transfer learning-based and meta-learning-based approaches. Transfer learning-based methods~\cite{dai2023pfemed,paul2021discriminative,yazdanpanah2022visual,DHeLZZGYZ22} aim to adapt to new medical data by training a highly generalized representation network on large-scale generic datasets. 
Meta-learning approaches~\cite{jiang2022multi,singh2021metamed,vo2022meta,khandelwal2020domain}. aim to acquire general knowledge, such as feature extraction capabilities, across a plethora of diverse tasks. Subsequently, the models rapidly iterate to achieve better results on small-sample data in the target domain.
These approaches focus on enhancing the generalization of the feature extraction network for obtaining more accurate representations of target domain features, while they fail to address the problem of learning the accurate category distribution with limited data, which directly leads to the suboptimal performance of the classifier.

\begin{figure*}[thb]
	\centering
	\includegraphics[width=1.0\linewidth]{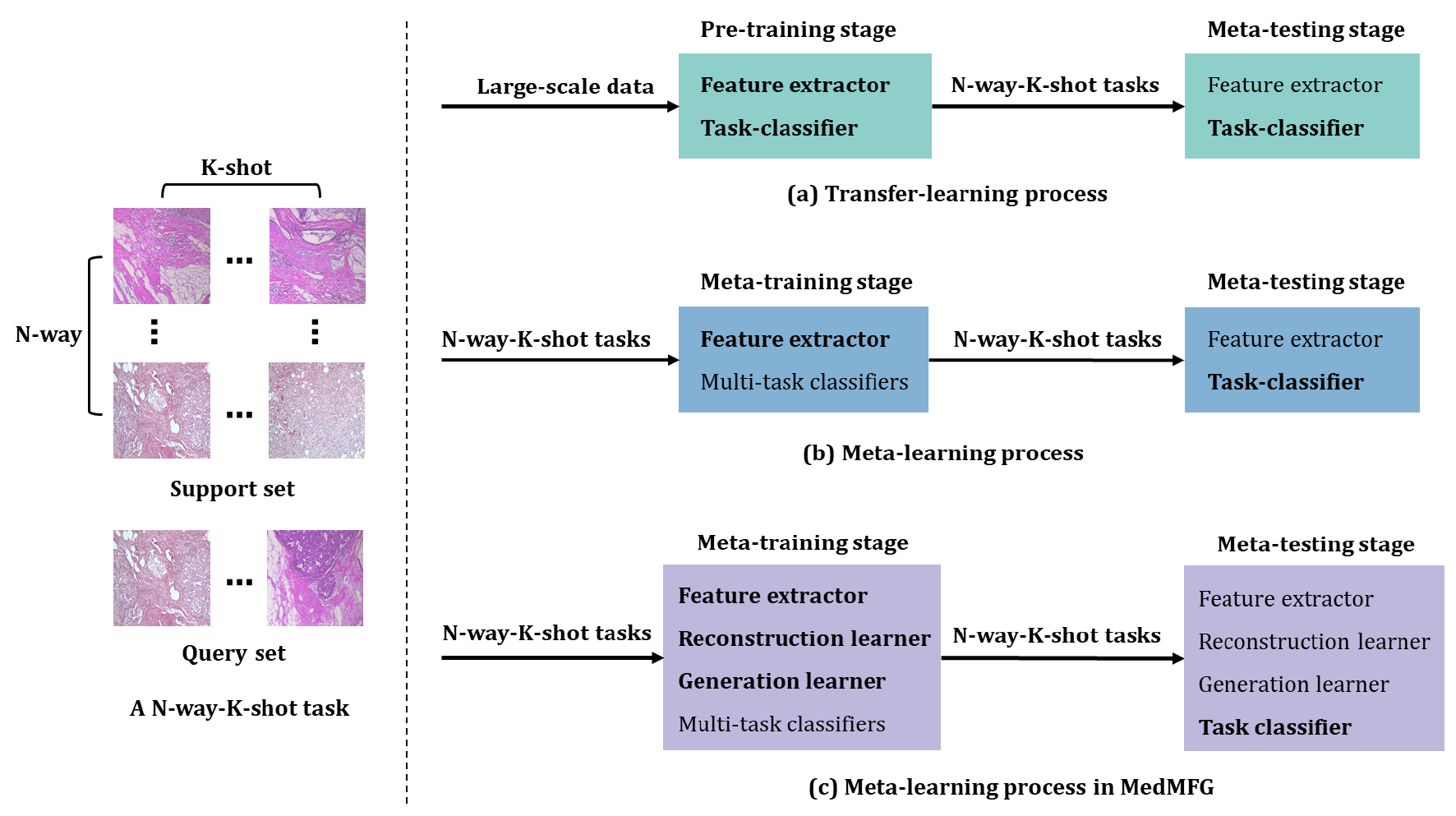}
	\caption{Comparison of training and testing process between transfer learning (a) and meta-learning (b). (c) is the process of our MedMFG. The bold parts in the stage boxes refer to the modules that are involved in training, while the others refer to the parts that are frozen. An example of an N-way-K-shot task is shown on the left.}
	\label{Fig:process}
\end{figure*}

To address this issue, we propose MedMFG, a plug-and-play feature generation method that focuses on augmenting sufficient data with a few prototype features. We propose augmenting the feature data by generating additional samples from the limited prototype, inspired by the variational autoencoder (VAE)~\cite{FotiKSC22, DanksY21}. This augmentation and process further enhance classifier training. Moreover, we incorporate multiple reconstruction methods before augmentation, motivated by attention mechanisms for feature representation~\cite{zhao2020exploring,vaswani2017attention}. In contrast to existing approaches that completely rely on backbones like ResNet-50, our reconstruction method employs a few simple transformer layers. These layers can be seamlessly incorporated after diverse pre-trained models to amplify the discriminative information of prototype features.

 Concretely, our method comprises three modules: a self-construction feature module, an inter-construction feature module, and a variational sample generation module. 
Firstly, the self-construction feature module utilizes self-attention mechanisms to assign higher weights to the discriminative information of prototype features. 
Subsequently, the inter-construction feature module re-expresses the information related to the test samples in the prototypes through cross-attention mechanisms.
Finally, the variational sample generation module generates a large number of effective feature samples with category-specific characteristics. These augmented features alleviate the distribution bias arising from limited prototype data. Finally, the generated features, combined with the prototype features, are utilized to train the classifier, significantly improving its generalization performance.

In summary, this paper has the following contributions:
\begin{itemize}
\item We introduce MedMFG, a flexible and lightweight plug-and-play method for generating features in few-shot medical image classification. It effectively addresses distribution bias stemming from overfitting by seamlessly integrating with diverse backbones and baselines.
\item We propose the self-construction and inter-construction modules to increase the weights of important information in the prototype features. Additionally, we introduce a variational sample generation module to augment these prototypes into suffcient class-discriminative features.
\item The efficacy and generalization of our approach are demonstrated through extensive experiments on CDFSL (cross-domain from natural to medical images) and FHIST (cross-domain across different lesions in medical images) benchmarks. Furthermore, fusion experiments illustrate multiple backbones and baselines can benefit from MedMFG.
\end{itemize}

\begin{figure*}[thb]
	\centering
	\includegraphics[width=1.0\linewidth]{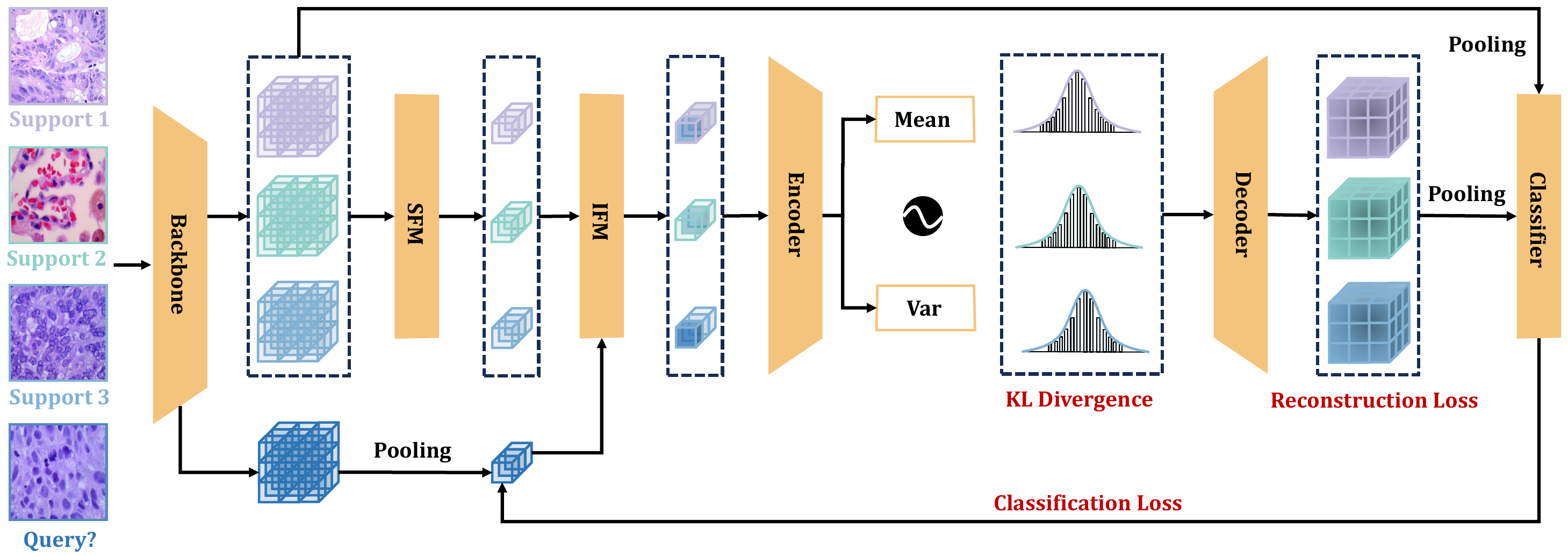}
	\caption{The framework of MedMFG is composed of the self-construction feature module (SFM), the inter-construction feature module (IFM), and the variational sample generation module (VSGM) which includes the encoder and decoder components.}
	\label{Fig:pipline}
\end{figure*}

\section{Preliminary and Problem Setting}\label{pre}
Few-shot learning (FSL)~\cite{LuGYZZ23} aims to recognize samples from unseen classes with limited training data, which is expected to reduce labeling costs and achieve a low-cost and quick model deployment.
Annotated samples are often hard to obtain in many specialized tasks, making FSL highly applicable in domains like fine-grained species recognition~\cite{tang2020revisiting}, industrial defect classification~\cite{cao2023effective}, and medical diagnosis~\cite{tang2021recurrent, singh2021metamed}. 
To simulate real-world scenarios, FSL datasets are typically split into a pre-training dataset and a testing dataset. Notably, the model has no prior exposure to the classes present in the testing dataset during training. Therefore, FSL requires online training of classifiers for new tasks with a few images. 

Specifically, there are two main differences between FSL and general classification. Firstly, the train set and test set in FSL have no overlapping classes. Specifically, FSL models are pretrained on the base dataset $ \mathcal{T}_{base}$ with base classes $ \mathcal{C}_{base}$, and they need to predict novel classes $ \mathcal{C}_{novel}$ on the novel dataset $\mathcal{T}_{novel}$, where ${C}_{base}\cap {C}_{novel} = \oslash$. Secondly, FSL utilizes a meta-task with the $N$-way-$K$-shot approach during both the training and testing phases, as depicted in the left part of Fig.~\ref{Fig:process}. In each meta-test ($i^{th}$ meta-test), the support sets $ \mathcal{S}_{i}$ and the query set $ \mathcal{Q}_{i}$ are randomly drawn from the $\mathcal{T}_{novel}$. Specifically, $N_{i}$ categories are randomly selected from $\mathcal{C}_{novel}$, and $K$ images per category are used as $\mathcal{S}_{i}$.
Obviously, the limited availability of $ \mathcal{S}_{i}$ often leads to severe overfitting of the classifiers, causing difficulties in accurately predicting the categories of $ \mathcal{Q}_{i}$.

In few-shot medical image classification, we adhere to the setting rules of classical few-shot learning (FSL). 
Apart from this, there is a notable difference in our experiments, as the domain gap between $ \mathcal{T}_{base}$ and $ \mathcal{T}_{novel}$ is more pronounced. For instance, in the CDFSL benchmark~\cite{guo2020broader}, $\mathcal{T}_{base}$ comprises natural images, whereas $\mathcal{T}_{novel}$ consists of two different modalities of medical images. 
This larger modality and the semantic gap in few-shot medical image classification make it difficult for pre-trained models to accurately represent the target classes in medical images. The inaccurately represented features exacerbate distribution bias when handling new class distributions. 
Therefore, obtaining accurate feature representation through re-representation and acquiring precise class distribution from limited data are key to improving the performance of few-shot medical image classification.

\section{Related Work}\label{pre}

\subsection{Transfer learning for FSL}
Transfer learning-based methods~\cite{rizve2021exploring,guo2023rankdnn,chen2019closer} primarily aim to enhance the generalization ability of the representation model on cross-domain data. As illustrated in Fig.~\ref{Fig:process}(a), these methods learn a universal image representation capability on the pre-training data $\mathcal{T}_{base}$. During the meta-testing phase, the feature extraction part of the model is typically frozen, and the classifier is trained online using $\mathcal{S}_{i}$, allowing predictions to be made on $\mathcal{Q}_{i}$.
Although transfer learning-based representation learning methods have been extensively researched and applied in traditional few-shot learning tasks, medical images present distinct challenges including diverse modalities, closer inter-class similarity, and the difficulty of directly transferring representation abilities from natural images. Consequently, our approach focuses on improving the performance of medical few-shot learning by directly increasing the number of training samples for medical target categories.

\subsection{Meta learning for FSL}
 Meta-learning-based methods~\cite{finn2017model,nichol2018reptile} strive to equip machines with the ability to learn how to learn. The objective is to achieve a universal classification capability through meta-training, enabling rapid adaptation to new tasks, as depicted in Fig.~\ref{Fig:process}(b). During the training process, meta-learning methods iterate tasks using the same $N$-way-$K$-shot setup as in testing. This iterative process results in a feature extractor that possesses robust generalization ability.
 Although meta-learning methods have shown promising performance in classical few-shot learning tasks, where they learn representation~\cite{abs-1803-02999, AntoniouES19, SongGYCPT20, CollinsMOS22}, classification~\cite{KaoCC22}, or optimization~\cite{AndrychowiczDCH16}, they often prove inadequate when faced with more complex domains like medicine and industry. Drawing inspiration from meta-learning approaches, we incorporate meta-training strategies, as in Fig.~\ref{Fig:process}(c). Specifically, in each training episode, we deliberately define visible and unseen samples to augment the model’s capability to fully consider information from unseen samples and facilitate feature interactions.

\section{Method}\label{med}
 \subsection{Method overview}
We propose MedMFG, as in Fig.~\ref{Fig:pipline}, consisting of three modules: the self-construction feature module (SFM), the inter-construction feature module (IFM), and the variational sample generation module (VSGM). 
After feature extraction, SFM and IFM enhance the class characteristics of the prototype features. Then, VSGM can generate sufficient class-discriminative samples for each prototype. These augmented samples are trained together with the prototype features to improve the generalization ability and classification accuracy of the classifier. Clearly, the training and utilization of MedMFG are independent of the feature extraction backbone, meaning that MedMFG can flexibly and widely integrate with various backbones or existing representation learning methods.

During the training phase, MedMFG adopts a meta-training approach. A training set $T_{Rj}$ is sampled, with $S_{Rj}$ serving as the support set and $Q_{Rj}$ as the query set. 
We follow exactly the same pre-training method as in other work to get the feature extraction backbone~\cite{2017Prototypical,2016Matching}.
Then, freezing the backbone, we train MedMFG using classification loss, KL divergence, and reconstruction loss.

In the meta-test phase, the backbone and MedMFG are both frozen. We only train classifiers online for each task. After feature extraction, SFM, and IFM, we obtain the support features $\hat{s}^{i}_{k}$ with updated weights from the original support images, as well as the query features $\hat{q}_{j}$ form the query image.
Next, we generate $n$ new samples for each support feature by VSGM. In the $N$-way-$K$-shot setup, we obtain $N$-way-$(k \times n)$ shot representative samples. Subsequently, we train the classifier using the generated new features along with the prototype features $N \times k \times (n+1)$ to classify $\hat{q}_{j}$.

\subsection{Model structure and loss function}

\begin{figure}[htb]
	\centering
	\includegraphics[width=0.80\linewidth]{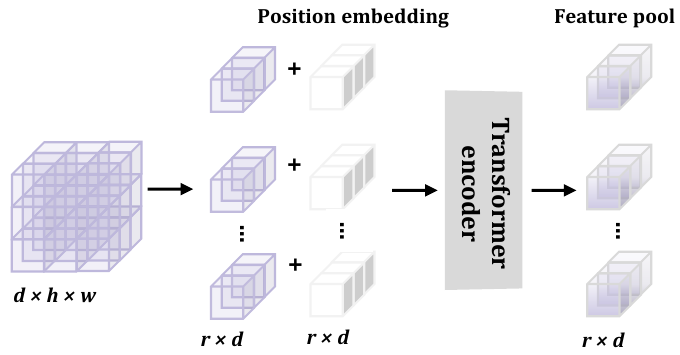}
	\caption{The self-construction feature module (SFM).}
	\label{Fig:SFM}
\end{figure}

\noindent{\bf{SFM:}} SFM increases the weight of category-specific information in prototype features through self-attention mechanisms. As depicted in Fig.~\ref{Fig:SFM}, the support samples $x_{i}$ are fed into the embedding module $f_{\theta}$, resulting in the prototype features $\hat{x}_{i}= f_{\theta}(x_i)\in\mathbb{R}^{d\times h\times w}$, where $d$ denotes the number of channels, and $h$ and $w$ represent the height and width of the features, respectively. In SFM, the inputs are $z_{i}=x_{i}+E_{pos}$, where $E_{pos}$ employs sinusoidal position encoding. The output of SFM is computed using the standard self-attention operation in the Transformer Encoder~\cite{dosovitskiy2020image}, performing multi-attention calculations using Eq. (\ref{self_attention})

\begin{equation}\label{self_attention}
{Attention}(Q,K,V) = {softmax}(\frac{{QK^{T}}}{\sqrt{d_{k}}}){V}.
\end{equation}

Thus, we obtain the re-characterized features $\hat{z}_{i}$ as:
\begin{equation}\label{SFM_result}
\hat{z}_{i} =  {Attention}(z_{i}W^{Q}_{\phi},z_{i}W^{K}_{\phi},z_{i}W^{V}_{\phi}),\hat{z}_{i}\in \mathbb{R}^{r\times d},
\end{equation}
where $W^{Q}_{\phi},W^{k}_{\phi},W^{V}_{\phi}$ are the learnable weights with $d\times d$ size. Finally, $z^{i}$ is calculated continually by a layer normalization and an MLP.
\begin{equation}\label{Sresult}
\hat{z}_{i} = MLP(LN(z_{i}+\hat{z}_{i})).
\end{equation}

\begin{figure}[thb]
	\centering
	\includegraphics[width=0.90\linewidth]{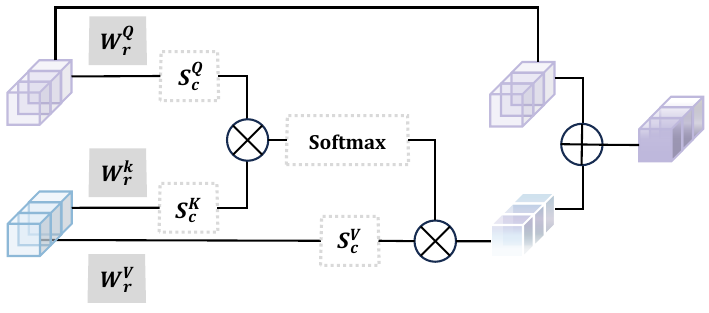}
	\caption{The inter-construction feature module (IFM).}
	\label{Fig:IFM}
\end{figure}

\noindent{\bf{IFM:}} IFM enhances the weight of information related to the test samples in prototype features through interactive attention mechanisms. After we obtain the self-reconstructed feature $\hat{z}_{i}$ of the support set, as well as the pooled query prototype feature ${q}_{i}$, we construct the IFM module shown as Fig.~\ref{Fig:IFM} to embed the attention weights between ${q}_{i}$ and $\hat{z}_{i}$ into the features of $\hat{z}_{i}$, resulting in the new support feature $\dot{z}_{i}$:
\begin{equation}\label{IFM_result}
\dot{z}_{i} =  {Attention}(\hat{z}_{i}W^{Q}_{r},{q}_{i}W^{K}_{r},{q}_{i}W^{V}_{r}),\dot{z}_{i}\in \mathbb{R}^{r\times d},
\end{equation}
where $Attention$ follows Eq.(\ref{self_attention}). Therefore, in order to better learn the feature interaction capability of the IFM module, we adopt a meta-training approach to better simulate the embedding of unknown features with known features.

\noindent{\bf{VSGM:}} VSGM utilizes an MLP-based VAE to
perform variational augmentation on the input features and
train a more generalized classifier. As in Fig.~\ref{Fig:pipline}, the VAE consists of two parts: an encoder $E(x,\theta_{e})$ which map the input feature $\dot{z}_{i}$ to a latent code $\tilde{z}$, and a decoder $G(\tilde{z},\theta_{g})$, which reconstructs $z_{i}$ from $\dot{z}_{i}$, where $\theta_{e}$ and $\theta_{g}$ are weight parameters of $E$ and $G$. 
Same as the traditional VAE~\cite{0Auto}, we approximate the true posterior distribution $p(z|X)$ with another distribution $q(z|X)$. Thus, the KL scatter between the two distributions is:
\begin{equation}\label{KL}
{KL(q(z|X)||p(z|X))} = \int_{Z}q(Z|X)\log \frac{q(z|X)}{(p(z|X)}.
\end{equation}

\begin{table*}[thpb]
\caption{Comparison between MedMFG and other state-of-the-arts of 5-way few-shot accuracies on FHIST benchmark with ResNet-18.}
 \centering
 \label{tab:FHIST benckmark}
 \scalebox{1.10}{
\begin{tabular}{lccccccccc}
\bottomrule
\multirow{2}{*}{{Method}} &\multicolumn{3}{c}{{CRC-TP$\rightarrow$ NCT}} &\multicolumn{3}{c}{{CRC-TP$\rightarrow$ LC25000}} &\multicolumn{3}{c}{{CRC-TP$\rightarrow$ BreakHis}}\\
\cline{2-10}
&1-shot &5-shot &10-shot &1-shot &5-shot &10-shot &1-shot &5-shot &10-shot\\
\hline
ProtoNet~\cite{2017Prototypical}
 &$58.1$ &$72.3$ &$74.4$ &$58.6$
&$68.5$ &$70.8$ &$32.0$ &$40.8$ &$44.0$\\
MetaOpt~\cite{2019Meta}
&$48.6$ &$64.4$ &$70.0$ &$49.2$
&$72.9$ &$78.5$ &$28.3$ &$42.5$ &$49.1$\\
SimpleShot~\cite{2019SimpleShot}
 &$63.2$ &$76.1$ &$78.4$ &$62.1$
&$73.8$ &$76.6$ &$36.2$ &$47.5$ &$51.8$\\
Distill~\cite{2020Rethinking}
 &$61.9$ &$77.3$ &$80.9$ &$63.2$
&$75.9$ &$79.5$ &$36.2$ &$50.2$ &$55.7$\\
Finetune~\cite{2019A}
&$64.7$ &$81.4$ &$85.0$ &$62.1$
&$79.2$ &$83.7$ &$36.4$ &$56.9$ &$66.7$\\

{\bf MedMFG}&{$\textbf{69.2}$} &{$\textbf{85.2}$} &{$\textbf{87.7}$}
&{$\textbf{65.3}$}
&{$\textbf{82.0}$}
&{$\textbf{84.4}$}
&{$\textbf{39.2}$}
&{$\textbf{59.9}$}
&{$\textbf{67.6}$}\\

\bottomrule

\end{tabular}}

 \label{tab:comparison on FHIST}
\end{table*}

\noindent{\bf{Loss function:}} We design a three-part loss function to optimize MedMFG: the KL scatter, the reconstruction loss, and the classification loss. 
To diversify the samples generated by VSGM, we set the prior distribution of $\tilde{z}$ to a centered isotropic multivariate Gaussian: $p(\tilde{z}) = N(0,I)$.
For the posterior distribution, we assume it to be a multivariate Gaussian with diagonal covariance:
$q({z}_{i}|X^{i}) = N(\mu^{i},\sigma^{i})$. The KL divergence$KL(q({z}_{i}|X^{i})||p(\tilde{z}))$ can be calculated by Eq. (\ref{KL}):
\begin{equation}\label{Loss_KL}
L_{KL} = \int_{z}q({z}_{i}|X^{i})\log \frac{q({z}_{i}|X^{i})}{p(\tilde{z}|X^{i})}.
\end{equation}
To drive VSGM to generate high-quality features, we design the reconstruction feature loss:
\begin{equation}\label{Loss_con}
L_{cons} = ||z_{i}-\dot{z}_{i}||^2.
\end{equation}
After obtaining augmented samples $z_{i}$, for each meta-training task $R_{j}$, we train an online classifier $W_{R_{j}}$ to predict the class of the query $q_{i}$, resulting in a classification loss:
\begin{equation}\label{Loss_clas}
L_{cls} = L_{(cross-entropy)}(W_{Rj}(q_{i}),y_{i}),
\end{equation}
where $y_{i}$ refers to the true label of $q_{i}$.
The classification loss is crucial as it can help the SFM and IFM maintain the class properties when reconstructing features, and the features generated by VSGM do not deviate from the original category boundaries.
Thus, the overall loss function is a weighted combination of the aforementioned terms:
\begin{equation}\label{Loss}
L = L_{cons} + \alpha L_{KL} + \beta L_{cls}.
\end{equation}
The performance is experimentally verified to be optimal for $\alpha = 0.0001$ and $\beta = 1$.

\begin{table*}[thpb]
\caption{Comparison between MedMFG and other state-of-the-arts of 5-way few-shot accuracies on miniImageNet$\rightarrow$ChestX/ISIC.}
 \centering
 \label{tab:CDFSL benckmark}
 \scalebox{1.10}{
\begin{tabular}{lcccccc}
\bottomrule
\multirow{2}{*}{{Method}} &\multicolumn{3}{c}{{miniImageNet$\rightarrow$ChestX}} &\multicolumn{3}{c}{{miniImageNet$\rightarrow$ISIC}} \\
\cline{2-4}\cline{5-7}
&5-shot &20-shot &50-shot &5-shot &20-shot &50-shot\\
\hline
MatchingNet~\cite{2016Matching} &$22.40$ &$23.61$ &$22.12$ &$36.74$ &$45.72$  &$54.58$ \\
MAML~\cite{finn2017model} &$23.48$ &$27.53$ &$--$ &$40.13$ &$52.36$  &$--$ \\
ProtoNet~\cite{2017Prototypical} &$24.05$ &$28.21$ &$29.32$ &$39.57$ &$49.50$  &$51.99$ \\
FWT~\cite{tseng2020cross} &$21.26$ &$23.23$ &$23.01$ &$30.40$ &$32.01$  &$33.17$ \\
FT~\cite{tseng2020cross} &$24.28$ &$--$ &$--$ &$40.87$ &$--$  &$--$ \\
FT-All~\cite{guo2020broader}&$25.97$ &$31.32$ &$35.49$ &$48.11$ &$59.31$  &{$\textbf{66.48}$} \\
AFA~\cite{wang2021cross} &$24.32$ &$--$ &$--$ &$44.91$ &$--$  &$--$ \\
ConFeSS~\cite{das2022confess} &$27.09$ &$33.57$ &$39.02$ &$48.85$ &$60.10$  &$65.34$ \\
AF~\cite{hu2022adversarial} &$25.02$ &$--$ &$--$ &$46.01$ &$--$  &$--$ \\
RDC~\cite{li2022ranking} &$25.91$ &$--$ &$--$ &$41.28$ &$--$  &$--$ \\
{\bf MedMFG}&{$\textbf{29.83}$} &{$\textbf{34.00}$} &$\textbf{39.66}$
&{$\textbf{50.03}$}
&{$\textbf{60.52}$}&$65.74$\\

\bottomrule

\end{tabular}}

 \label{tab:comparison on CDFSL}
\end{table*}

\begin{table*}[ht]
\caption{Comparison experiments of MedMFG with multiple bakcbones and baselines of 5-way few-shot accuracies on miniImageNet$\rightarrow$ChestX/ISIC.}
 \centering
 \label{tab:fusion MedMFG}
 \scalebox{1.10}{
\begin{tabular}{lccccccc}
\bottomrule
\multirow{2}{*}{Method} &\multirow{2}{*}{Backbone} &\multicolumn{3}{c}{miniImageNet $\rightarrow$ ChestX} &\multicolumn{3}{c}{miniImageNet $\rightarrow$ ISIC} \\
\cline{3-5}\cline{6-8}
& &5-shot &20-shot &50-shot &5-shot &20-shot &50-shot\\
\hline
Baseline-1~\cite{2017Prototypical}
&Conv-4 &$24.04$ &$26.23$ &$28.49$ &$35.32$
&$40.17$ &$44.69$\\
{\bf{+MedMFG}} &Conv-4
&{$27.50_{{\color{red}+3.46}}$} &{$29.97_{{\color{red}+3.74}}$}
&{$31.50_{{\color{red}+3.01}}$}
&{$38.25_{{\color{red}+2.93}}$}
&{$44.52_{{\color{red}+4.35}}$}
&{$48.24_{{\color{red}+3.55}}$}\\
\hline
Baseline-1~\cite{2017Prototypical}
&ResNet-10 &$24.05$ &$28.21$ &$29.32$ &$39.57$
&$49.50$ &$51.99$\\
{\bf{+MedMFG}}&ResNet-10 & {$28.30_{{\color{red}+4.25}}$} &{$31.17_{{\color{red}+2.96}}$}
&{$33.00_{{\color{red}+3.68}}$}
&{$44.52_{{\color{red}+4.95}}$}
&{$53.13_{{\color{red}+3.63}}$}
&{$55.50_{{\color{red}+3.51}}$}\\
\hline
Baseline-1~\cite{2017Prototypical}
&ResNet-18 &$23.13$ &$29.02$ &$33.20$ &$42.83$
&$53.48$ &$58.51$\\
{\bf{+MedMFG}} &ResNet-18 
&{$26.94_{{\color{red}+3.81}}$} &{$32.22_{{\color{red}+3.20}}$}
&{$39.55_{{\color{red}+6.35}}$}
&{$49.00_{{\color{red}+6.17}}$}
&{$56.71_{{\color{red}+3.23}}$}
&{$62.11_{{\color{red}+3.60}}$}\\
\hline
Baseline-2~\cite{2016Matching}
&ResNet-10 &$22.40$ &$23.61$ &$24.12$ &$36.74$
&$45.72$ &$54.58$\\
{\bf{+MedMFG}} &ResNet-10
&{$25.98_{{\color{red}+3.58}}$} &{$26.86_{{\color{red}+3.25}}$}
&{$27.11_{{\color{red}+2.99}}$}
&{$39.80_{{\color{red}+3.06}}$}
&{$48.99_{{\color{red}+3.27}}$}
&{$57.60_{{\color{red}+3.02}}$}\\
\hline
Baseline-3~\cite{2017Learning}
&ResNet-10 &$22.96$ &$26.63$ &$28.45$ &$39.41$
&$41.77$ &$49.32$\\
{\bf{+MedMFG}}  &ResNet-10
&{$26.00_{{\color{red}+3.04}}$} &{$30.17_{{\color{red}+3.54}}$}
&{$32.05_{{\color{red}+3.60}}$}
&{$42.67_{{\color{red}+3.26}}$}
&{$45.21_{{\color{red}+3.44}}$}
&{$52.44_{{\color{red}+3.12}}$}\\
\bottomrule

\end{tabular}}

 \label{tab:fusion on CDFSL}
\end{table*}

\section{Experiments}
\label{sec:experiments}

\subsection{Experimental setup}
\noindent{\bf{Datasets:}} We validate our method on two benchmarks: CDFSL~\cite{guo2020broader} and FHIST~\cite{shakeri2022fhist}, following the standard few-shot setting as in previous works~\cite{parnami2022learning,he2022attribute,guo2023rankdnn}. Specifically, on CDFSL, we regard miniImageNet~\cite{2016Matching}, a dataset excerpted from ImageNet, as the base set while ISIC~\cite{tschandl2018ham10000}, a dermoscopic image dataset and chestX~\cite{wang2017chestx}, a frontal-view X-ray images dataset as the novel set respectively.
On FHIST, we train MedMFG on CRC-TP~\cite{javed2020multiplex} (colorectal tissues images) and test it on LC25000 (colorectal and lung tissues images)~\cite{borkowski2019lung}, BreakHis (breast tissues images)~\cite{spanhol2015dataset}, and NCT-CRC-HE-100k (colorectal tissues images)~\cite{pataki2022huncrc}. 
It is important to note that we follow the inductive setting, ensuring no information leakage among different queries.

\noindent{\bf{Learnable parameters:}} MedMFG is designed to be lightweight and portable. Among the three modules, SFM consists of one transformer block with 1.66M parameters, IFM consists of another transformer block with 1.24M parameters, and the VAE module has 0.79M parameters. In total, MedMFG has 3.69M parameters. To ensure a fair comparison, we use the same feature extractor as the comparison methods.

\noindent{\bf{Implementation details:}} During the meta-training phase, we follow the 5-way-1-shot approach for training. In the meta-testing phase, after extracting the features of support and novel data, we apply MedMFG to amplify each support feature into $5$ samples. We report the accuracy on standard N-way K-shot settings with $15$ query samples per class. All testing experiments are randomly sampled over 600 episodes. The learning rate of MedMFG is set to 0.00001, and it is optimized with Adam.

\subsection{Result analysis}
\noindent{\bf{Comparison to state-of-the-Art (SOTA) methods:}} We compare our method with the SOTA methods on CDFSL and FHIST benchmarks. On FHIST, as shown in Tab.~\ref{tab:comparison on FHIST}, MedMFG achieves the new state-of-the-art in all three migration settings. For instance, on CRC-TP$\rightarrow$ NCT, MedMFG outperforms Finetune~\cite{2019A} by $4.5\%$, $3.8\%$, and $2.7\%$ on 5-way 1-shot, 5-shot, and 10-shot settings, respectively. In other scenarios like CRC-TP$\rightarrow$ LC25000 and CRC-TP$\rightarrow$ BreakHis, our method surpasses the current optimal results by $3.2\%$, $2.8\%$, $0.7\%$, $2.8\%$, $3.0\%$, and $0.9\%$. It is worth noting that Finetune~\cite{2019A} requires retraining during testing, whereas our approach is completely frozen during testing without any additional consumption of time and space. These results demonstrate the superiority and practicality of MedMFG in a wide range of medical domain migrations.

On CDFSL, as shown in Tab.~\ref{tab:comparison on CDFSL}, MedMFG achieves a new SOTA on most of the results. For example, on miniImageNet$\rightarrow$ChestX and ISIC, MedMFG outperforms ConFess~\cite{das2022confess} by $2.74\%$, $0.43\%$, $1.18\%$, and $0.42\%$ on 5-way 5-shot and 20-shot settings. In the 50-shot setting, our method reaches the current optimum. These results highlight the strong generalization ability of MedMFG.

In summary, whether it involves cross-domain transfer from natural images to medical images or cross-category transfer on medical images, MedMFG demonstrates stronger generalization capabilities. This proves the effectiveness of the feature design and augmentation methods in medical images, providing a novel approach to address the challenges posed by the limited availability of medical data.

\noindent{\bf{Fusion experiments:}} To demonstrate the generalizability and flexibility of our method, we extract features using multiple backbones and baselines, and then fuse MedMFG amplification samples. 
As shown in Tab.~\ref{tab:fusion on CDFSL}, we validate the fusion results of MedMFG with three feature extractors, Conv-4~\cite{KrizhevskySH12}, ResNet-10~\cite{HeZRS16}, and ResNet-18~\cite{HeZRS16}, as well as three methods ~\cite{2017Prototypical,2016Matching,2017Learning}. MedMFG consistently improves the performance by more than 2.9\% across all settings. Particularly, it surpasses the baseline by 6\% in the two results with ResNet-18.
Notably, it exhibits superior performance even with fewer support instances, indicating that the generated feature samples accurately capture class characteristics. These results highlight the versatility of MedMFG, which is not restricted to specific datasets or feature extractors. As a result, various works based on different backbones can benefit from our approach.

\begin{table}[ht]
\caption{Ablation experiments of modules and the number of MLP layers in VAE.}
 \centering
 \label{tab:Ablation MedVAE}
 \scalebox{1.0}{
\begin{tabular}{lcccccc}
\bottomrule
\multirow{2}{*}{Method} &\multirow{2}{*}{SFM} &\multirow{2}{*}{IFM} &\multirow{2}{*}{Layers}  &\multicolumn{3}{c}{\bf CRC-TP$\rightarrow$ NCT} \\
 \cline{5-7}
& & & &1-shot &5-shot &10-shot\\
\hline
Baseline~\cite{2017Prototypical} &$\times$ &$\times$  &$\times$ &58.1 &72.3 &74.4 \\
MedMFG-1&$\checkmark$ &$\times$ &$\times$&58.7 &73.6 &75.7 \\
MedMFG-2 &$\checkmark$ &$\checkmark$ &$\times$  &59.0 &75.3 &79.0\\
MedMFG-3 &$\checkmark$ &$\checkmark$ &one  &67.3 &84.0 &84.3\\
MedMFG-4 &$\checkmark$  &$\times$ &two &63.1 &77.0 &76.3\\
MedMFG-5 &$\times$  &$\checkmark$&two &66.5 &83.9 &83.7\\
MedMFG-6 &$\checkmark$  &$\checkmark$&three &69.0 &\textbf{85.7} &86.9\\
MedMFG-7 &$\times$  &$\times$ &two &60.0 &75.6 &76.1\\
MedMFG  &$\checkmark$  &$\checkmark$ &two &\textbf {69.2} & 85.2 &\textbf {87.7}\\

\bottomrule
\end{tabular}}

\end{table}

\noindent{\bf{Ablation studies:}} In Tab.~\ref{tab:Ablation MedVAE}, we analyze the effectiveness of the three modules in MedMFG: SFM, IFM, and VSGM. Additionally, we compare the performance with varying numbers of layers in the VAE. The results show that using SFM or IFM alone does not lead to significant improvements, especially in the 1-shot setting. This is because, without the feature augmentation module, a single sample fails to adequately represent the entire distribution of a class, resulting in a substantial bias. However, the last three rows clearly demonstrate the crucial role of SFM and IFM in conjunction with feature augmentation. The absence of SFM or IFM results in a performance decrease of more than 2.5\%. This indicates that feature reconstruction effectively highlights class characteristics and reduces intra-class distances, contributing to improved performance.
Moreover, when comparing MedMFG-3 with MedMFG, it can be observed that the VAE with a two-layer MLP structure better simulates the characteristics of real samples. However, blindly increasing the number of MLP layers provides minimal benefits while adding more parameters and increasing training complexity. Additionally, in the 20-shot setting, MedMFG-6 performs worse than MedMFG, suggesting that excessive feature expression can limit feature diversity. Therefore, we ultimately choose the VAE composed of a two-layer MLP as the final structure.

\begin{figure*}[th]
	\centering
	\includegraphics[width=1.0\linewidth]{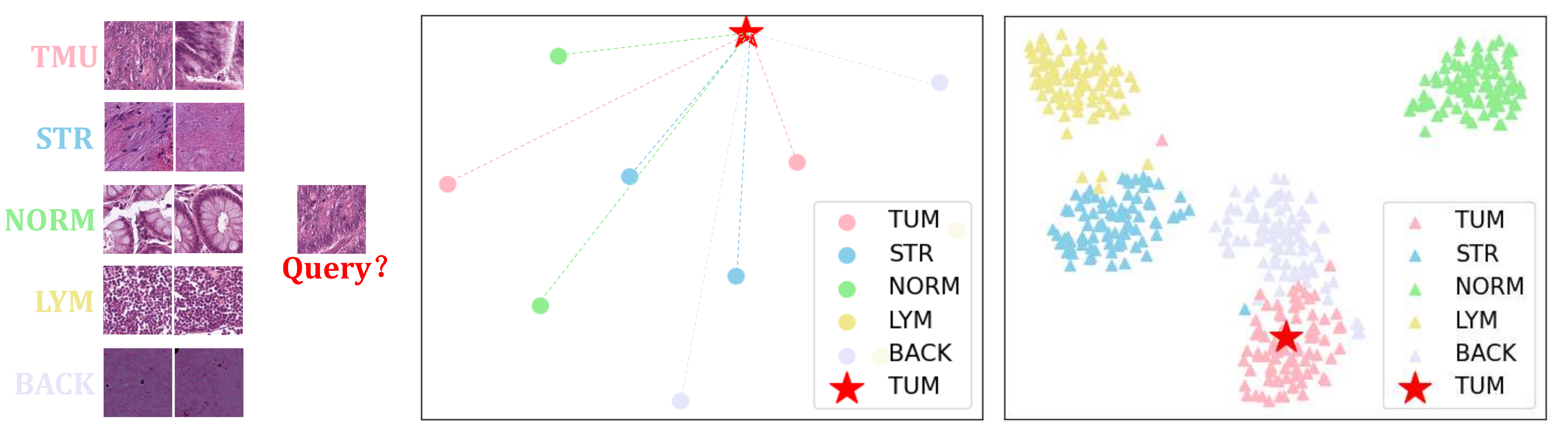}
	\caption{For a 5-way-2-shot task, we visualize the feature descriptors generated by the feature backbone (Left) and after amplification by MedMFG (Right) with t-SNE.}
	\label{Fig:tsne}
\end{figure*}

\subsection{Visualization}

We assert that MedMFG is capable of generating representative samples to enhance classification results. To visualize this, we present the original features generated by the feature extractor and the augmented features of MedMFG on the NCT dataset in Fig.~\ref{Fig:tsne}. In this example, the true label of the query is TMU. However, due to the fine-grained nature of the support classes, the difference in distance between the query and each support feature is small, resulting in overfitting of the linear regression classifier and misclassification errors. Nevertheless, after variational feature generation by MedMFG to adjust the distribution, the query is accurately included within the distribution boundary of TMU.

\section{Conclusion}
\noindent{\bf{Summary:}} In this paper, to enhance the generalization ability of the classifier in few-shot medical image classification, we propose MedMFG. It is a plug-and-play method that can augment sufficient class-discriminative features from a small number of samples.
Through comparison experiments, we demonstrate the superiority of our method by achieving new SOTA in multiple benchmarks. Furthermore, fusion experiments showcase the flexibility and applicability of our method.

\noindent{\bf{Limitations:}} By comparing the baseline with MedMFG-1, as well as MedMFG-4 and MedMFG-7 in Tab.~\ref{tab:Ablation MedVAE}, it is apparent that employing the self-attention mechanism alone yields limited improvement over the baseline. In contrast, when combined with VSGM, the self-attention mechanism demonstrates promising performance. The cross-attention mechanism exhibits similar characteristics. We will investigate the generalizability of this phenomenon through additional experiments. Furthermore, we aim to explore additional medical modalities like CT and MRI, enhancing the practicality of MedMFG.

\bibliographystyle{IEEEtran}
\bibliography{reference.bib}

\end{document}